%% file: main.tex
\pdfoutput=1

\documentclass[11pt]{article}

\usepackage[]{acl}

\usepackage{times}
\usepackage{latexsym}
\usepackage{booktabs}
\usepackage{multirow}
\usepackage{adjustbox}
\usepackage{setspace}

\usepackage{enumitem}
\usepackage{times}
\usepackage{latexsym}
\usepackage{diagbox}

\usepackage[T1]{fontenc}
\usepackage[utf8]{inputenc}
\usepackage{hyperref}

\usepackage{microtype}
\usepackage{colortbl} 

\usepackage{inconsolata}
\usepackage{multirow}
\usepackage{graphicx}
\usepackage{amsmath}
\usepackage{url}   
\usepackage{nicefrac} 
\usepackage{longtable}
\usepackage{bbm}

\usepackage{amssymb}
\usepackage{xcolor}
\usepackage{booktabs}
\usepackage{arydshln}
\usepackage{subcaption}
\usepackage{needspace}
\usepackage{hyperref}

\definecolor{purple_f}{HTML}{B9B8EF}
\definecolor{purple_b}{HTML}{D4D5FF}
\definecolor{green_f}{HTML}{385723}
\definecolor{green_b}{HTML}{E2F0D9}
\usepackage[most]{tcolorbox}
\tcbset{on line, 
    boxsep=1.0pt, left=0pt, right=0pt,top=0pt,bottom=0pt,
    colframe=white, arc=1pt
}

\usepackage[most]{tcolorbox}

\usepackage[T1]{fontenc}

\usepackage[utf8]{inputenc}

\usepackage{microtype}

\usepackage{inconsolata}

\usepackage{graphicx}

\definecolor{arrowgreen}{RGB}{34, 139, 34}
\definecolor{arrowred}{RGB}{178, 34, 34}

\definecolor{gray}{HTML}{D3D3D3} 
\definecolor{green}{HTML}{E2F0D9}

\usepackage{xspace}
\newcommand{\weight}{{\tt[WEIGHT]}\xspace}

\title{One for All: Update Parameterized Knowledge Across Multiple Models with Once Edit}

\author{
    Weitao Ma$^{1}$\footnotemark[1] \quad Xiyuan Du$^{1}$\thanks{$\quad$ means Equal Contribution} \quad Xiaocheng Feng$^{1,2}$\thanks{$\quad$ means Corresponding Author} \quad Lei Huang$^{1}$ \quad Yichong Huang$^{1}$ \\ \textbf{Huiyi Zhang$^{1}$ \quad Xiaoliang Yang$^{1}$ \quad Baohang Li$^{1}$ \quad Xiachong Feng$^{3}$ \quad Ting Liu$^{1}$ \quad Bing Qin$^{1,2}$} \\
    $^{1}$ Harbin Institute of Technology \\
    $^{2}$ Peng Cheng Laboratory \\
    $^{3}$ The University of Hong Kong \\
    \texttt{\{wtma,xcfeng\}@ir.hit.edu.cn}
}

\begin{document}
\maketitle

\begin{abstract}
Large language models (LLMs) encode vast world knowledge but struggle to stay up-to-date, often leading to errors and hallucinations.
Knowledge editing offers an efficient alternative to retraining, enabling targeted modifications by updating specific model parameters.
However, existing methods primarily focus on individual models, posing challenges in efficiently updating multiple models and adapting to new models.
To address this, we propose \textsc{OnceEdit}, a novel ensemble-based approach that employs a plug-in model as the editing module, enabling stable knowledge updates across multiple models.
Building on the model ensemble, \textsc{OnceEdit} introduces two key mechanisms to enhance its effectiveness.
First, we introduce a \textit{dynamic weight mechanism} through a \weight token for distinguishing between edit-related and non-edit-related instances, ensuring the appropriate utilization of knowledge from integrated models.
Second, we incorporate an \textit{ensemble enhancement mechanism} to mitigate the excessive reliance on the central model inherent in the model ensemble technique, making it more suitable for knowledge editing.
Extensive experiments on diverse LLMs demonstrate that \textsc{OnceEdit} consistently outperforms existing methods while achieving superior editing efficiency.
Further analysis confirms its adaptability and stability in multi-model editing scenarios. Our code will be available.
\end{abstract}

\section{Introduction}{\label{sec:intro}}
\input{section/Introduction}

\section{Preliminaries}{\label{sec:preliminary}}
\input{section/Preliminaries}

\section{Methodology}\label{sec:methods}
\input{section/Methods}

\section{Experiments}\label{sec:experiments}
\input{section/Experiments}

\section{Further Analysis and Ablation Study}\label{sec:analysis}
\input{section/Analysis}

\section{Related Work}\label{sec:related_work}
\input{section/Related_work}

\section{Conclusion}\label{sec:conclu}
\input{section/Conclusion}

\section*{Limitations}\label{sec:limitations}
\input{section/Limitations}

\section*{Acknowledgements}\label{sec:Acknowledgments}
\input{section/Acknowledgements}

\bibliography{custom}

\appendix
\include{section/Appendix}

\end{document}

%% file: section/Introduction.tex
Large language models \citep{achiam2023gpt, jiang2023mistral, meta2024introducing} have demonstrated remarkable performance in various downstream tasks by scaling in both parameters and training data, thereby capturing extensive world knowledge during pretraining \citep{wang2024knowledge, feng2023trends}.
However, as real-world information undergoes dynamic changes, the internal parameterized knowledge of LLMs gradually becomes outdated, resulting in errors and hallucinations \citep{huang2023survey, zhang2023large, zhong2024investigating}, hindering the practical application of LLMs. 
Currently, efforts to mitigate hallucinations mainly focus on two aspects: faithfulness \citep{huang2024advancing, huang2025improving} and factuality \citep{li2024dawn}. 
Among these, knowledge editing has emerged as a promising approach for improving the factual accuracy of LLMs by directly modifying their internal knowledge. 
Rather than resorting to costly retraining, knowledge editing provides an efficient and practical ways to update a model’s knowledge \citep{yao2023editing, zhang2024comprehensive}.
These techniques enable the integration of growing knowledge into the models by allowing precise updates through the targeted parameters modification \citep{li2024consecutive}.

\begin{figure}[t]
\centering
\includegraphics[width=1.0\linewidth]{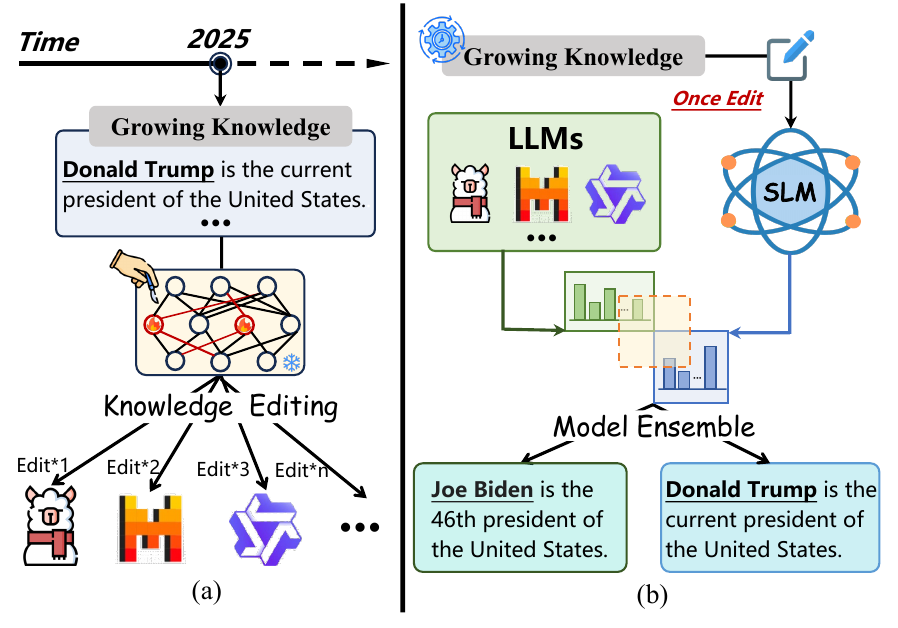}
\caption{The comparison of traditional knowledge editing and \textsc{OnceEdit} for multi-model updates. (a) Traditional methods require separate edits for each model, while (b) \textsc{OnceEdit} updates all models with a single edit via model ensemble.
}
\label{fig:comparison_knowledge_editing}
\end{figure}

In recent years, various knowledge editing methods for LLMs have been proposed, leveraging techniques such as meta-learning \citep{mitchell2021fast, tan2023massive}, locate-then-edit strategies \citep{meng2022locating, meng2022mass}, and memory-based approaches \citep{mitchell2022memory, hartvigsen2024aging} to update model knowledge while preserving unrelated information. 
However, existing methods primarily focus on modifying a single model, which makes them unsuitable for complex scenarios requiring the simultaneous update of multiple models. 
Additionally, these methods exhibit significant sensitivity to hyperparameter settings, leading to considerable inconsistency in editing effectiveness across models, which limits their scalability and adaptability to new models.

To address these challenges, we introduce \textsc{OnceEdit}, which modifies a unified lightweight plug-in model and employs a heterogeneous model ensemble for knowledge transfer across multiple models, thereby enabling seamless and stable knowledge editing, as shown in Figure \ref{fig:comparison_knowledge_editing}. 
However, model ensemble methods are not directly applicable to editing scenarios, and \textsc{OnceEdit} introduces two improvement mechanisms to align more closely with knowledge editing tasks. 
Firstly, traditional ensemble methods fuse the knowledge of the plug-in model and the LLM using fixed ensemble weights, making them unsuitable for knowledge editing, where new knowledge should be updated without affecting unrelated information.
To this end, \textsc{OnceEdit} introduces a \textit{dynamic weight mechanism} using a special \weight token, which predicts weight allocation for each instance, ensuring the effective utilization of knowledge from integrated models.
Secondly, model ensembles often suffer from the inherent bias of the central large model, where its knowledge dominates the ensemble results compared to the plug-in model. 
To counter this, we propose an \textit{ensemble enhancement mechanism} that incorporates two strategies: \textit{search-space zero initialization} and \textit{target augmentation}. 
By starting the decoding search with a zero vector instead of the central model's distribution, and emphasizing high-probability tokens from the fused distribution, these strategies ensure that the decoding is driven by the fused knowledge, improving both the precision and generalization of the edited knowledge.

We conduct extensive experiments on Llama2-7B, Mistral-7B-v0.1, and GPT-J-6B using the ZsRE and Counterfact datasets to compare the performance of \textsc{OnceEdit} against seven popular knowledge editing methods. 
Experimental results demonstrate that \textsc{OnceEdit} consistently outperforms other methods in both teacher-forced and validation generation evaluation settings, which better align with realistic scenarios, while also requiring the fewest editing interventions.
Additionally, we quantitatively analyze the editing time of each method, showing that \textsc{OnceEdit} incurs the lowest editing overhead in multi-model knowledge editing scenarios. 
Furthermore, we extend our evaluation to more and larger models, such as Qwen2.5-7B, Llama3-70B, etc., to further validate the adaptability and stability of \textsc{OnceEdit}.

%% file: section/Preliminaries.tex
In this section, we introduce knowledge editing as the core task of our study and model ensemble as the underlying technique supporting our methods.
\subsection{Knowledge Editing}
\label{KE_def}
Knowledge editing is an effective technique for updating LLMs with new knowledge. 
Given a target model that is parametrized by $\theta$ and a new edited set $S_E$, the goal of knowledge editing is to update the model so that it correctly responds to the edits while maintaining its unrelated knowledge. 
The knowledge editing function, denoted as $\text{KE}(\theta, S_E)$, represents the process of modifying the model $\theta$ based on the edited knowledge set $S_E$. The editing process can be expressed as follows:
\begin{equation}
    \theta^{\prime} \leftarrow \text{KE}(\theta, S_E),
\end{equation}
let $f_\theta(\cdot)$ represent the original mapping function of the model $\theta$. The expected output of the edited model $\theta^{\prime}$ is defined as follows:
\begin{equation}
    f_{\theta^{\prime}}(x) = 
    \begin{cases}
    y_e & \text{if }x \in I_{edit},\\
    f_{\theta}(x) & \text{otherwise}.
    \end{cases}
\end{equation}
Here, $I_{edit}$ represents the set of instances within the editing scope of the edits in $S_E$. In addition to $S_E$, $I_{edit}$ may also include knowledge-related input, such as re-phrased versions of the edit input.

Following previous research \citep{wang2024wise, zhang2024comprehensive}, an ideal knowledge editing method should ensure that the edited model meets three key properties: \textbf{Reliability}, \textbf{Generality}, and \textbf{Locality}. 
These properties collectively ensure that the edited model maintains correctness on targeted updates, generalizes appropriately, and preserves unaffected knowledge. 
Details about these three properties can be found in Appendix \ref{appendix:eval_KE}.

\begin{figure*}[t]
\centering
\includegraphics[clip, width=1.0\linewidth]{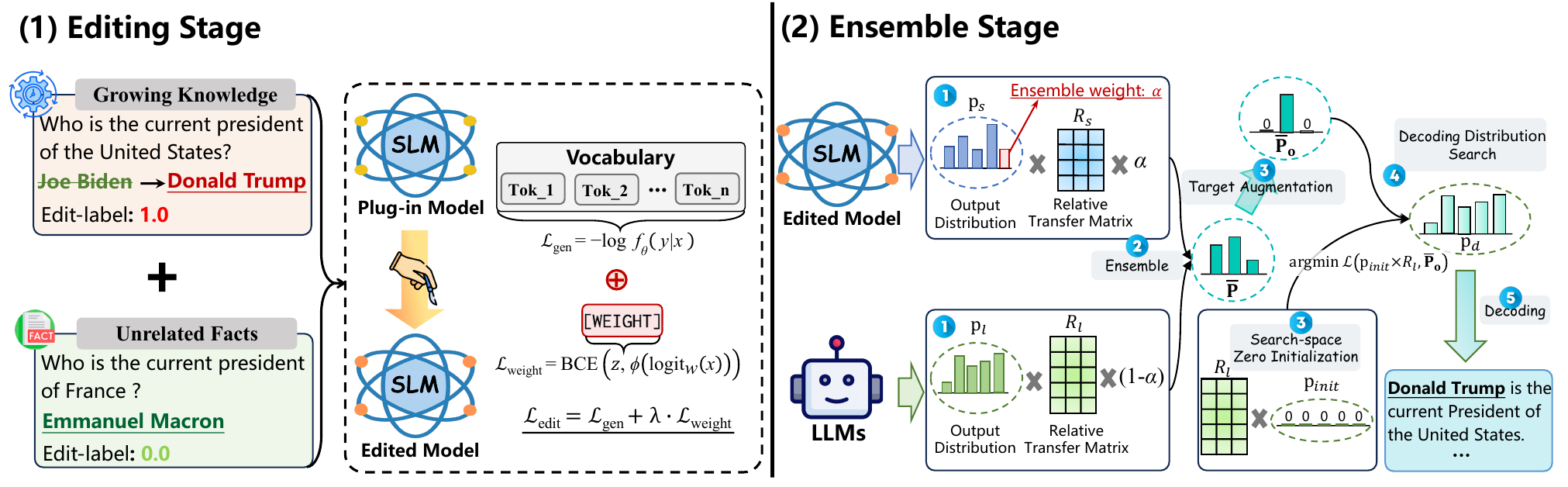}
\caption{Overview of \textsc{OnceEdit}, which consists of two stages. In the editing stage, \textsc{OnceEdit} applies knowledge edits to a lightweight model while introducing \weight to learn the ensemble weights (\S\ref{Editing_stage}). 
In the ensemble stage, the edited model is integrated with LLMs to achieve multi-model knowledge updating (\S\ref{Ensemble_stage}).
}
\label{fig:methods}
\end{figure*}

\subsection{Model Ensemble}
\label{ME_def}
Existing knowledge editing methods primarily focus on single models, making it difficult to efficiently adapt across models. 
This challenge motivates us to propose a model ensemble framework, where a small plug-in model serves as an edited module to reliably modify various LLMs.

The model ensemble techniques integrate the output distributions from multiple models to achieve an optimal result.
Specifically, the aggregated output probability, denoted as $\overline{\mathbf{P}}$, is calculated as the sum of the weighted probabilities of each model, where $\alpha_i$ represents the weight assigned to the $i$-th model and $\mathbf{p}_i$ is the output probability distribution from the $i$-th model. The ensemble process can be expressed as follows:
\begin{equation} 
\overline{\mathbf{P}} = \sum_{i=1}^{N} \alpha_i \times \mathbf{p}_i.
\end{equation}
However, when the candidate models to be integrated are heterogeneous, additional steps are needed to align their vocabularies before effective integration. 
For instance, in the classic heterogeneous model ensemble method, DEEPEN \citep{huang2024enabling}, the procedure involves selecting a set of common tokens shared across models to serve as anchor words. 
The distance between other words and each anchor word is then computed, resulting in a relative transfer matrix. 
This matrix is used to map each model's output probability into the relative representation space, facilitating the integration of their probability distributions. 
The process can be expressed as follows:
\begin{equation}
\overline{\mathbf{P}} = \sum_{i=1}^{N} \alpha_i \times (\mathbf{p}_i \times R_i),
\end{equation}
where $R_i \in \mathbb{R}^{|V_i|\times |A|}$ represents the relative transfer matrix of the $i$-th model. Here, $|V_i|$ denotes the corresponding vocabulary size of the $i$-th model and $|A|$ represents the number of anchor words.

%% file: section/Methods.tex
In this section, we introduce \textsc{OnceEdit} based on DEEPEN \citep{huang2024enabling}, an effective method for integrated heterogeneous models.
The overall process of \textsc{OnceEdit} comprises two stages: \textit{Editing Stage} and \textit{Ensemble Stage}, as shown in Figure \ref{fig:methods}.
In the editing stage, \textsc{OnceEdit} selects a lightweight plug-in model as the editing module, which is updated using knowledge editing techniques to incorporate new information  (\S\ref{Editing_stage}). 
To enhance DEEPEN’s static weight allocation, \textsc{OnceEdit} introduces the \textbf{Dynamic Weight} echanism, which enables instance-level weight adjustment.
In the ensemble stage, the edited model is integrated with multiple LLMs for knowledge updating (\S\ref{Ensemble_stage}). To better align DEEPEN with knowledge editing, \textsc{OnceEdit} incorporates the \textbf{Ensemble Enhancement} mechanism, leveraging \textit{Search-space Zero Initialization} and \textit{Target Augmentation} to stabilize the new knowledge transfer.

\subsection{Editing Stage}
\label{Editing_stage}
To efficiently facilitate multi-model knowledge updates, editing the plug-in model plays a crucial and foundational role. 
In this context, \textsc{OnceEdit} employs a simple but effective full fine-tuning strategy to update the knowledge within the plug-in model. 
This approach is particularly well-suited since the plug-in model is relatively small, keeping the associated computational cost manageable. 
The following outlines the training objectives:
\begin{align}\label{eq:sft}
    & \mathcal{L}_{\text{gen}}(\theta) = 
    -\mathbb{E}_{(x, y)\in S_E}
    \left[
      \log f_\theta \left( y \,| x \right)
    \right].
\end{align}
However, full fine-tuning often results in significant degradation of the model's original knowledge. 
To address this issue, \textsc{OnceEdit} introduces a \textbf{Dynamic Weighting} mechanism that adaptively adjusts the contribution of each model during the ensemble stage based on the given input.
Specifically, we introduce a special token, \weight, into the vocabulary of the plug-in model. 
This token helps distinguish between knowledge that requires modification and knowledge that should remain unaffected. 
Consequently, for edit-related inputs, the plug-in model is assigned a higher weight, whereas for non-edit-related inputs, the LLMs dominate.

To effectively train \weight, it is essential not only to fine-tune the model on the target knowledge modifications but also to introduce a set of unrelated knowledge as a reference group to guide the model in distinguishing between edit-related and non-edit-related knowledge.
The training objective of the token is formulated as follows:
\begin{align}\label{eq:weight}
    & \mathcal{L}_{\text{weight}}(\theta) = \mathbb{E}_{(x, z)}
      \text{BCE} \left(z, \phi(\text{logit}_w(x) \right),
\end{align}
where $z$ denotes the edit-label of the input, where instances related to edits are assigned a value of 1, and instances not related to edits are assigned a value of 0. Additionally, $\text{logit}_w(x)$ represents the logits at the position of the \weight after encoding the input $x$. In this context, $\phi(\cdot)$ and $\text{BCE}(\cdot)$ denote the sigmoid function and the binary cross-entropy loss function, respectively.

Finally, we adopt a multi-task learning approach to jointly train the plug-in model. The overall training objective of the editing stage is formulated as:
\begin{align}\label{eq:edit}
    & \mathcal{L}_{\text{edit}}(\theta) = \mathcal{L}_{\text{gen}}(\theta) +  \lambda \cdot \mathcal{L}_{\text{weight}}(\theta),
\end{align}
where $\lambda$ is a hyperparameter that balances the learning contributions of the two tasks.

\subsection{Ensemble Stage}
\label{Ensemble_stage}
During the ensemble stage, as previously described in \S\ref{ME_def}, we select the common words shared between the plug-in model and the LLMs as anchor words and calculate the corresponding relative transfer matrix. 
At each decoding step, the ensemble models use the corresponding relative transfer matrices to map the output distribution into the relative space, where it is then fused with weighted contributions.
The aggregated distribution is then obtained by combining the outputs, with the weight provided by \weight:
\begin{equation}
\overline{\mathbf{P}} = \alpha \times (\mathbf{p}_s \times R_s) + (1-\alpha) \times (\mathbf{p}_l \times R_l),
\end{equation}
where $\mathbf{p}_s$ and $\mathbf{p}_l$ represent the output distributions of the plug-in model and the LLMs. $R_s$ and $R_l$ are the relative transfer matrices for these models. $\alpha = \phi(\text{logit}_w(x))$ is the ensemble weight derived from the plug-in model. 

Once the aggregated distribution is obtained, we use the LLM as the decoding model. 
Following the DEEPEN framework, we employ gradient descent to search for an optimal output distribution within the vocabulary space of the LLM, ensuring that the aggregated distribution is accurately represented. 
The process is formalized as follows:
\begin{equation}
\label{eq:fused}
\begin{aligned}
\textbf{p}_{d} = \mathop{\arg\min} \mathcal{L} (\textbf{p}_{init} \times R_l, \ \overline{\mathbf{P}}),
\end{aligned}
\end{equation}
where $\textbf{p}_{init}$ and $\textbf{p}_{d}$ denote the initial search distribution and the final decoding distribution, both in the absolute representation space of the LLMs.

DEEPEN originally initializes the search using the LLM’s output distribution $\textbf{p}_{init}=\textbf{p}_l$, treating the aggregated distribution as a perturbation to the LLM’s original output. 
This approach is effective for traditional model ensemble, where the integrated models produce similar outputs, allowing for minor corrections for LLM’s behaviors.
However, this method can lead to the central model becoming biased, resulting in the ensemble's output being overly reliant on the LLM's knowledge. 
In the context of knowledge editing, where the plug-in model and the LLM often exhibit significant distributional differences, using the LLM’s original distribution for initialization may fail to effectively capture the newly injected knowledge.

Based on the above analysis, we propose an \textbf{Ensemble Enhancement} mechanism including two strategies: \textit{Search-space Zero Initialization} and \textit{Target Augmentation} to better align the decoding distribution with the aggregated distribution. These strategies work together to strengthen the search process, as described below:
\begin{equation}
\label{eq:zero_search}
\begin{aligned}
 \textbf{p}_{init}&= \operatorname{zeros\_like}(\textbf{p}_{l}), \\
\end{aligned}
\end{equation}
\begin{equation}
\label{eq:target_aug}
\begin{aligned}
\overline{\mathbf{P_o}} &=
\begin{cases}
1, & i = \arg\max_j \overline{\mathbf{P}}_j, \\
0, & \text{otherwise},
\end{cases}
\end{aligned}
\end{equation}
where $\operatorname{zeros\_like}(\cdot)$ is a function that creates a vector with the same shape as the input, but with all elements set to 0.

\input{section/tables/main_table_1000_main}

These two strategies work as follows: the first, Search-space Zero Initialization, initializes the search space with a zero vector. The second, Target Augmentation, converts the aggregated distribution into a one-hot vector. 
Together, these strategies help the final decoding distribution better capture the partial order relations in the aggregated distribution, ultimately leading to a more effective representation of the new knowledge.

%% file: section/tables/main_table_1000_main.tex
\begin{table*}[ht]
\centering
\resizebox{\linewidth}{!}{
\begin{tabular}{lcccccccccccccc}
    \toprule
    \multirow{2}{*}{\textbf{Method}}
    & \multicolumn{4}{c}{\textbf{Llama2-7B}} & \multicolumn{4}{c}{\textbf{Mistral-7B-v0.1}} & \multicolumn{4}{c}{\textbf{GPT-J-6B}} & \multirow{2}{*}{\textbf{Score}$\uparrow$} & \multirow{2}{*}{\textbf{Freq.}$\downarrow$} \\
    \cmidrule(lr){2-5} \cmidrule(lr){6-9} \cmidrule(lr){10-13}
    & Rel.$\uparrow$ & Gen.$\uparrow$ & Loc.$\uparrow$ & Avg.$\uparrow$ 
    & Rel.$\uparrow$ & Gen.$\uparrow$ & Loc.$\uparrow$ & Avg.$\uparrow$ 
    & Rel.$\uparrow$ & Gen.$\uparrow$ & Loc.$\uparrow$ & Avg.$\uparrow$
    &  & \\
    \midrule
          \multicolumn{15}{c}{\textbf{ZsRE}} \\
    \midrule
    FT-L    &0.34  &0.21  &0.13 &0.23 &0.55 &0.41 &0.54 &0.50 &0.11 &0.10 &0.48 &0.23 &0.32 &3 \\
    MEND    &0.00  &0.00  &0.00 &0.00 &0.00 &0.00 &0.00 &0.00 &0.00 &0.00 &0.00 &0.00 &0.00 &3 \\
    ROME    &0.07  &0.06  &0.01 &0.05 &0.01 &0.01 &0.01 &0.01 &0.01 &0.01 &0.00 &0.01 &0.02 &3 \\
    MEMIT   &0.78  &0.76  &0.53 &0.69 &\underline{0.91} &\textbf{0.89} &0.50 &\underline{0.77} &\textbf{0.98} &\textbf{0.92} &0.76 &\textbf{0.89} &0.78 &3 \\
    DEFER   &0.63  &0.58  &\underline{0.62} &0.61 &0.37 &0.36 &\textbf{1.00} &0.58 &0.34 &0.32 &0.85 &0.50 &0.56 &3 \\
    WISE    &\underline{0.84}  &\underline{0.78}  &\textbf{0.99} &\underline{0.87} &0.68 &0.64 &\underline{0.99} &\underline{0.77} &0.76 &0.68 &\textbf{1.00} &0.81 &\underline{0.82} &3  \\
    \midrule 
    \rowcolor{blue!15} 
    \textsc{OnceEdit}  & \textbf{0.99} &\textbf{0.92}  &\textbf{0.99} &\textbf{0.97} &\textbf{0.95} &\underline{0.88} &0.98 &\textbf{0.93} &\underline{0.84} &\underline{0.76} &\underline{0.99} &\underline{0.87} &\textbf{0.92} &1 \\
    \midrule
        \multicolumn{15}{c}{\textbf{Counterfact}} \\
    \midrule
    FT-L    &0.26  &0.01  &0.18 &0.15 &0.41 &0.05 &\textbf{0.99} &0.48 &0.71 &0.09 &0.07 &0.30 &0.31 &3\\
    MEND    &0.00  &0.00  &0.00 &0.00 &0.00 &0.00 &0.00 &0.00 &0.00 &0.00 &0.00 &0.00 &0.00 &3 \\
    ROME    &0.07  &0.04  &0.05 &0.05 &0.01 &0.02 &0.00 &0.01 &0.00 &0.00 &0.00 &0.00 &0.02 &3 \\
    MEMIT   &0.95  &0.51  &0.23 &0.56 &\underline{0.78} &0.43 &0.26 &0.49 &\textbf{0.99} &0.20 &\textbf{0.90} &\underline{0.70} &0.58 &3 \\
    DEFER   &\underline{0.98}  &\textbf{0.88}  &0.34 &\underline{0.74} &0.47 &\underline{0.53} &\underline{0.79} &\underline{0.60} &\underline{0.94} &\textbf{0.84} &0.15 &0.64 &\underline{0.66} &3 \\
    WISE    &0.74  &0.33  &\underline{0.38} &0.48 &0.67 &0.24 &0.35 &0.42 &0.37 &0.08 &0.37 &0.27 &0.39 &3 \\
    \midrule 
    \rowcolor{blue!15} 
    \textsc{OnceEdit}  &\textbf{0.99} &\underline{0.81} &\textbf{0.62} &\textbf{0.81} &\textbf{0.94} &\textbf{0.76} & 0.62 &\textbf{0.77} &\underline{0.94} &\underline{0.76} &\underline{0.53} &\textbf{0.74} &\textbf{0.72} &1 \\
    \bottomrule 
\end{tabular}
}
\caption{Experimental results on ZsRE and Counterfact under teacher-forced setting. \textbf{Bold} and \underline{underline} numbers indicate the best and second performance among evaluated methods. Score represents the average output of the three models (Rel., Gen., and Loc.), while Freq. indicates the total number of edits required to update these models.}
\label{tab:main_results}
\end{table*}

%% file: section/Experiments.tex
\subsection{Experimental Setups}
\paragraph{Datasets} Building on previous works \citep{meng2022locating, yao2023editing}, we conduct our experiments using two widely-used model editing datasets: ZsRE \citep{levy2017zero} and Counterfact \citep{meng2022locating}. ZsRE is a context-free question-answering dataset, and we adopt the dataset split following \citet{zhang2024comprehensive}. Counterfact is a counterfactual dataset in its completed form, which is employed to assess the impact of model editing techniques on entity-relation triples.

\input{section/tables/ablation_results}

\paragraph{Metrics} We evaluate all methods from three perspectives based on the EasyEdit \citep{wang2023easyedit}, as defined in \S\ref{KE_def}: Reliability (Rel.), Generality (Gen.), and Locality (Loc.), which are commonly used in prior works \citep{wang2024wise, hartvigsen2024aging}. 
The final score is the average accuracy across these three sets.

\paragraph{Baselines} We select seven trending baselines compared with \textsc{OnceEdit}, covering four distinct types of knowledge editing methods: 
1) \textit{Constrained fine-tuning}: \textbf{FT-L} \citep{meng2022locating}, focuses on fine-tuning a single layer's FFN with new knowledge while incorporating an additional KL divergence loss. 
2) \textit{Locate-then-edit}: \textbf{ROME} \citep{meng2022locating} and \textbf{MEMIT} \citep{meng2022mass}, employ causal tracing to identify model areas relevant to the desired edit, followed by targeted updates to the corresponding parameters.
3) \textit{Meta-learning}: \textbf{MEND} \citep{mitchell2021fast}, trains an external hyper-network to model the gradients produced by conventional fine-tuning.
4) \textit{Memory-based}: This category encompasses the \textbf{DEFER} \citep{hartvigsen2024aging}, \textbf{WISE} \citep{wang2024wise}, and \textbf{GRACE} \citep{hartvigsen2024aging} methods, all of which use dedicated memory to store and manage edited knowledge. 

\paragraph{Implementation Details} 
We conduct experiments on three popular models from prior research: Llama2-7B \cite{touvron2023llama}, Mistral-7B-v0.1 \cite{jiang2023mistral}, and GPT-J-6B \cite{wang2021gpt}. 
For the datasets, we sample 1,000 records from the evaluation sets of ZsRE and Counterfact under the batch editing setting, where the evaluation is conducted after all knowledge editing operations have been completed.
Meanwhile, we select Tiny-Llama \cite{zhang2024tinyllama} as the plug-in model for \textsc{OnceEdit}.
Additionally, we utilize EasyEdit for evaluation, incorporating two decoding strategies: teacher-forced and validation generation. 
For the main experiment, we apply both strategies, with the teacher-forced strategy being commonly employed in prior research and the validation generation strategy better reflecting real-world scenarios. 
Among the baselines, GRACE primarily reports results based on the validation generation strategy. 
Further details are provided in Appendix \ref{app:main_exp_details}.

\subsection{Main Results}
\label{main_results}
The main experimental results are shown in Table \ref{tab:main_results}. \textsc{OnceEdit} achieves the highest overall scores, surpassing the second-best methods by 14\% on ZsRE and 6\% on Counterfact.
Unlike traditional knowledge editing methods that require separate edits for each model, \textsc{OnceEdit} updates multiple models with a single edit, demonstrating its efficiency. 
Moreover, \textsc{OnceEdit} achieves top performance in all five settings except for a slightly lower score on GPT-J-6B with ZsRE, demonstrating its strong capability in single-model editing.

Additionally, the results indicate that other methods exhibit significant performance fluctuations across datasets and models. 
For instance, MEMIT performs well on GPT-J-6B with Counterfact but poorly on ZsRE (0.92 vs. 0.20), and FT-L shows exceptionally high locality (0.99) on Mistral-7B-v0.1 under Counterfact while underperforming on other models. 
In contrast, \textsc{OnceEdit} maintains stable and consistent results across all models and datasets.
Notably, due to the nature of Counterfact completions, all methods yield lower locality on it compared to ZsRE. Despite this, \textsc{OnceEdit} achieves comparable locality across different models.
Overall, the main experiment demonstrates that \textit{\textsc{OnceEdit} enables stable and effective knowledge editing across multiple models and datasets.}

\begin{figure}[tp]
\centering
\includegraphics[width=1.0\columnwidth]{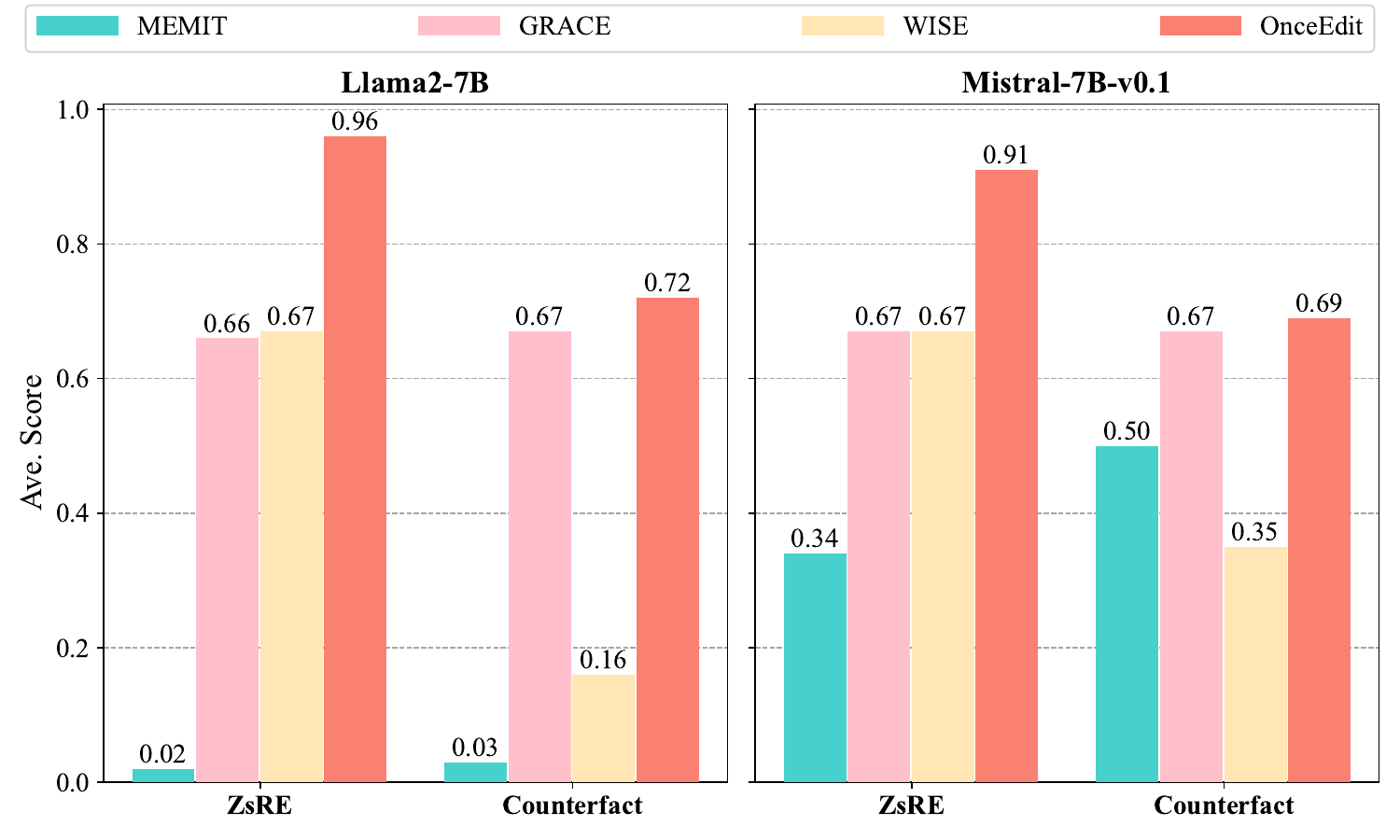} 
\caption{Experimental results on ZsRE and Counterfact under validation generation setting. Score represents the average output of each model (Rel., Gen., and Loc.).}
\label{fig:generation_results}
\end{figure}

Previous studies typically evaluate methods under teacher-forced conditions. 
However, the validation generation setting, which relies solely on output generation, more accurately reflects a model's ability to understand and apply the knowledge, and better aligns with real-world scenarios.
As a result, we choose to supplement our evaluation with this setting.
The experimental results, presented in Figure \ref{fig:generation_results}, highlight four methods that stand out in terms of performance. 
Notably, \textsc{OnceEdit} exhibits superior editing capabilities, outperforming all other methods. 
In contrast, MEMIT, which performs relatively well under teacher-forced, shows a significant drop in performance under the validation generation setting. 
Although GRACE achieves a strong overall score, it demonstrates poor generalization, a limitation that has also been observed in \citet{wang2024wise}.
More detailed results can be found in Table \ref{tab:generation_results}.

%% file: section/tables/ablation_results.tex
\begin{table*}[t]
\centering
\resizebox{\linewidth}{!}{
\begin{tabular}{lccccccccccccc}
    \toprule
    \multirow{2}{*}{\textbf{Method}}
    & \multicolumn{4}{c}{\textbf{Llama2-7B}} & \multicolumn{4}{c}{\textbf{Mistral-7B-v0.1}} & \multicolumn{4}{c}{\textbf{GPT-J-6B}} & \multirow{2}{*}{\textbf{Score}$\uparrow$} \\
    \cmidrule(lr){2-5} \cmidrule(lr){6-9} \cmidrule(lr){10-13}
    & Rel.$\uparrow$ & Gen.$\uparrow$ & Loc.$\uparrow$ & Avg.$\uparrow$ 
    & Rel.$\uparrow$ & Gen.$\uparrow$ & Loc.$\uparrow$ & Avg.$\uparrow$ 
    & Rel.$\uparrow$ & Gen.$\uparrow$ & Loc.$\uparrow$ & Avg.$\uparrow$
    & \\
    \midrule
          \multicolumn{14}{c}{\textbf{ZsRE}} \\
    \midrule
    DEEPEN &0.88 &0.81 &0.02 &0.57 &0.73 &0.67 &0.02 &0.47 &0.20 &0.19 &0.14 &0.18 &0.41 \\
    \hspace{0.3cm}+\textit{DW} &0.88 &\cellcolor{gray!80}0.77 &\cellcolor{green}0.99 &\cellcolor{green}0.88 &0.73 &\cellcolor{gray!80}0.65 &\cellcolor{green}0.95 &\cellcolor{green}0.78 &0.20 &0.19 &\cellcolor{green}0.99 &\cellcolor{green}0.46 &\cellcolor{green}0.71 \\
    \hspace{0.3cm}+\textit{DW}+\textit{EE}(Ours) &\cellcolor{green}0.99 &\cellcolor{green}0.89 &0.99 &\cellcolor{green}0.96 &\cellcolor{green}0.93 &\cellcolor{green}0.83 &\cellcolor{green}0.96 &\cellcolor{green}0.91 &\cellcolor{green}0.84 &\cellcolor{green}0.76 &0.99 &\cellcolor{green}0.86 &\cellcolor{green}0.91\\
    \midrule
        \multicolumn{14}{c}{\textbf{Counterfact}} \\
    \midrule
    DEEPEN &0.79 &0.60 &0.11 &0.50 &0.64 &0.51 &0.14 &0.43 &0.10 &0.06 &0.11 &0.09 &0.34 \\
    \hspace{0.3cm}+\textit{DW} &\cellcolor{green}0.81 &0.60 &\cellcolor{green}0.40 &\cellcolor{green}0.60 &0.64 &0.51 &\cellcolor{green}0.41 &\cellcolor{green}0.52 &\cellcolor{green}0.11 &\cellcolor{green}0.07 &\cellcolor{green}0.29 &\cellcolor{green}0.16 &\cellcolor{green}0.43  \\
    \hspace{0.3cm}+\textit{DW}+\textit{EE}(Ours) &\cellcolor{green}0.99 &\cellcolor{green}0.82 &\cellcolor{gray!80}0.36 &\cellcolor{green}0.72 &\cellcolor{green}0.94 &\cellcolor{green}0.76 &\cellcolor{gray!80}0.37 &\cellcolor{green}0.69 &\cellcolor{green}0.93 &\cellcolor{green}0.74 &\cellcolor{gray!80}0.23 &\cellcolor{green}0.63 &\cellcolor{green}0.68 \\
    \bottomrule 
\end{tabular}
}
\caption{Ablation study on the Dynamic Weight (DW) mechanism by \weight and the Ensemble Enhancement (EE) mechanism which includes \textit{Search-space Zero Initialization} and \textit{Target Augmentation}. \tcbox[colback=green]{Green} indicates improved performance compared to the previous row, while \tcbox[colback=gray!80]{gray} indicates a decline compared to the previous row.}
\label{tab:ablation_results}
\end{table*}

%% file: section/Analysis.tex
\subsection{Ablation Study}
In this section, we conduct a series of ablation studies to evaluate the effectiveness of \textsc{OnceEdit}'s components and the impact of hyperparameters. 

\input{section/tables/ablation_plugin_models}

\paragraph{Components Ablation.}  We examine the impact of the mechanisms introduced in \S\ref{sec:methods} on 1,000 edited instances under validation generation. 
The results are presented in Table \ref{tab:ablation_results}.
\textsc{OnceEdit} progressively enhances the overall editing performance of the three models by incorporating the dynamic weight mechanism and the ensemble enhancement mechanism into DEEPEN.
When applied to a single model on the datasets, the method with both mechanisms also achieves the best performance.

To evaluate the impact of these two mechanisms separately, we analyze their effects on model performance across different settings.
For the dynamic weight mechanism, results indicate that it significantly enhances locality while maintaining reliability and generality in most cases.
The only notable drawback is a minor generalization loss of 2\% to 4\% on Llama2-7B and Mistral-7B-v0.1 on ZsRE, which remains within an acceptable range.
Meanwhile, the ensemble enhancement mechanism proves to be highly beneficial, substantially improving reliability and generality across all settings in line with our design objectives. 
While it introduces a 4\% to 6\% reduction in locality for the three models on the Counterfact dataset, this trade-off is outweighed by gains in the other two aspects. 
Notably, the overall average score of each model on Counterfact increases by more than 12\%, making the slight decrease in locality an acceptable compromise.

\paragraph{Hyperparameters Ablation.} We conduct an ablation study on the balanced hyperparameter $\lambda$ for multi-task learning during the editing stage. 
Specifically, we select 200 edited instances and evaluate the performance under five different $\lambda$ settings. 
The experimental results, presented in Table \ref{tab:lamda_ablation}, indicate that $\lambda$ is relatively robust. 
With the exception of the extreme case where $\lambda =0$, all other settings yield good editing performance.

\paragraph{Plugin Models.}
In the main experiments, we selected Tiny-Llama as the plugin model, as it is one of the most widely used lightweight models. 
Moreover, in practical applications, OnceEdit demonstrates high adaptability, enabling seamless integration with various LLMs through a unified plugin model. 
This design supports flexible knowledge updates without requiring a switch between different plugin models.

However, to further explore the performance of alternative plugin models, we conducted additional experiments on the ZsRE dataset using Qwen2.5-1.5B as the plugin model, integrated with three different LLMs. 
We compared its overall knowledge editing performance against strong baselines, including FT-L, MEMIT, WISE, and DEFER, as presented in Table \ref{tab:ab_plugin}.

The results show that the \textsc{OnceEdit} method, when using Qwen2.5-1.5B as a plugin model, achieves a higher overall score than all baselines, further validating the effectiveness of OnceEdit. Its performance is only second to TinyLlama, which may be influenced by the post-editing performance of the plugin model itself. In practical applications, selecting a more stable small model (e.g., TinyLlama) as the plugin model can lead to more effective knowledge editing.

\input{section/tables/editing_time_cost_stats}

\input{section/tables/model_flops_tokens}

\subsection{Cost of Editing and Ensemble}
In this section, we quantitatively evaluate the editing time required for each method under identical hardware conditions and dataset scales.

The editing time statistics are summarized in Table \ref{tab:time_cost}.
MEMIT involves computing second-order momentum to ensure locality, while MEND requires training an additional hypernetwork using a training set. 
Due to the substantial computational cost of these two methods, their results are omitted from the table.
FT-L incurs lower time overhead for single-model editing as it only updates a specific layer of the model. Similarly, GRACE maintains a working memory for a specific layer, resulting in relatively low editing time as well.
However, as the number of models to be edited increases, the efficiency of \textsc{OnceEdit} becomes increasingly evident. 
When editing three models, the editing time for other methods is more than twice that of \textsc{OnceEdit}. 
This demonstrates that \textsc{OnceEdit} is highly efficient in multi-model knowledge editing scenarios, with its advantages becoming more pronounced as the number of models increases.

In addition, we analyze the migration cost of \textsc{OnceEdit} between LLMs, which mainly comes from the calculation of the relative transfer matrix of the integrated models. 
We combined DEEPEN to derive the FLOPS formula required to calculate the relative transfer matrices for the plug-in model and LLMs. 
The detailed derivation process is shown in the Appendix \ref{app:flops}.
We calculate the total FLOPS, denoted as $\text{TFLOPS}_{all}$, for the relative transfer matrix required for each pair of integrated models based on Equations \ref{eq:all_l} and \ref{eq:all_s}.
The results are summarized in Table \ref{tab:flops_cost}. 
Specifically, for the three models being edited, the cost of constructing the relative transfer matrices is equivalent to generating approximately 600 to 1000 tokens during forward propagation. 
This overhead is modest compared to the cost of performing another round of editing, demonstrating that the migration overhead introduced by our model integration is acceptable. 
% Overall, these quantitative results further confirm the high efficiency of \textsc{OnceEdit} in multi-model editing scenarios.

\input{section/tables/extend_model}

\subsection{Extending on More Models}
In the main experiment, we aim to provide an effective comparison with other popular knowledge editing methods by following previous studies and selecting three classic models.
Unlike other methods, which are often sensitive to hyperparameters, \textsc{OnceEdit} demonstrates strong adaptability and can be quickly generalized to new models. 
To further validate this, we selected four newer, larger, and more diverse models, including Llama3-8B, Mistral-7B-v0.3, Qwen2.5-7B, and Llama3-70B, and applied \textsc{OnceEdit} to edit 200 instances under the validation generation setting.
The results, as shown in Table \ref{tab:extend_results}, highlight \textsc{OnceEdit}'s ability to achieve effective and stable editing across all four models and two datasets, further confirming its high scalability.

%% file: section/tables/ablation_plugin_models.tex
\begin{table}[!t]
\centering
\resizebox{\linewidth}{!}{
\begin{tabular}{lcccc}
    \toprule
    \textbf{Methods}
    & Llama2-7B & Mistral-7B-v0.1 & GPT-J-6B & Score \\
    \midrule
    FT-L  &0.23  &0.50  &0.23 &0.32 \\
    MEMIT  &0.69  &0.77  &\textbf{0.89} &0.78  \\
    DEFER  &0.61  &0.58  &0.50 &0.56 \\
    WISE  &\underline{0.87}  &0.77  &0.81 &0.82 \\
    \midrule 
    \rowcolor{blue!15} 
    \textbf{\textsc{OE}(Tiny-Llama)} & \textbf{0.97} & \textbf{0.93} & \underline{0.87} & \textbf{0.92} \\
    \rowcolor{blue!15} 
    \textbf{\textsc{OE}(Qwen2.5-1.5B)} & 0.83 & 0.83 & \textbf{0.89} & \underline{0.85} \\
    \bottomrule 
\end{tabular}
}
\caption{Experimental results on the ZsRE dataset using different plugin models. \textbf{\textsc{OE}} represents the \textsc{OnceEdit} method. \textbf{Bold} and \underline{underline} numbers indicate the best and second performance among evaluated methods.}
\label{tab:ab_plugin}
\end{table}

%% file: section/tables/editing_time_cost_stats.tex
\begin{table}[!t]
    \centering
    \resizebox{\columnwidth}{!}{
    \begin{tabular}{lcccc}
    \toprule
         Methods &  Llama2-7B & Mistral-7B-v0.1 & GPT-J-6B & Total \\
         \midrule
         FT-L &  0.69& 0.71& 0.73 & 2.13 \\
         MEND &  -& -& -& - \\
         ROME &  2.29 & 3.39& 2.50& 8.18 \\
         MEMIT &  -& -& -& - \\
         GRACE &  0.65& 0.72& 0.84& 2.21 \\
         DEFER & 1.49& 1.47& 1.40& 4.36 \\
         WISE & 1.35& 1.33& 1.26& 3.94 \\
         \midrule 
         \rowcolor{blue!15} 
         \textbf{\textsc{OnceEdit}} &  1&  1& 1& 1 \\
         \bottomrule
    \end{tabular}
    }
    \caption{The editing times for each method are normalized relative to the time taken by \textsc{OnceEdit}, with '-' indicating a time that is more than 100 times longer. The term 'total' refers to the overall time required to edit all three models.}
    \label{tab:time_cost}
\end{table}

%% file: section/tables/model_flops_tokens.tex
\begin{table}[t!]
\centering
\resizebox{\columnwidth}{!}{
\begin{tabular}{lccc}
\toprule
     Models &  TFLOPS/token & $\text{TFLOPS}_{all}$ & Num \\
     \midrule
     Llama2-7B &  0.013& 12.638&  973 \\
     Mistral-7B-v0.1 &  0.014& 9.552&  683\\
     GPT-J-6B &  0.011& 9.739& 886\\
     \bottomrule
\end{tabular}
}
\caption{Analysis of the TFLOPS computational cost for model ensemble attributed to the relative representation matrix across three models.}
\label{tab:flops_cost}
\end{table}

%% file: section/tables/extend_model.tex
\begin{table}[ht]
\centering
\resizebox{\linewidth}{!}{
\begin{tabular}{lcccc}
    \toprule
    \textbf{Model}
    & Rel.$\uparrow$ & Gen.$\uparrow$ & Loc.$\uparrow$ & Avg.$\uparrow$ \\
    \midrule
          \multicolumn{5}{c}{\textbf{ZsRE}} \\
    \midrule
    Llama3-8B  &0.91  &0.90  &0.97 &0.93 \\
    Mistral-7B-v0.3  &0.93  &0.93  &0.95 &0.94  \\
    Qwen2.5-7B  &0.90  &0.90  &0.74 &0.85 \\
    Llama3-70B  &0.74  &0.75  &0.91 &0.80 \\
    \midrule
          \multicolumn{5}{c}{\textbf{Counterfact}} \\
    \midrule
    Llama3-8B  &0.99 &0.91 &0.16 &0.69 \\
    Mistral-7B-v0.3  &0.95 &0.87 &0.25 &0.69 \\
    Qwen2.5-7B  &0.98 &0.92 &0.18 &0.70 \\
    Llama3-70B  &0.79 &0.72 &0.18 &0.56 \\
    \bottomrule 
\end{tabular}
}
\caption{Extended experimental results on editing 200 instances across multiple models using \textsc{OnceEdit}.}
\label{tab:extend_results}
\end{table}

%% file: section/Related_work.tex
\paragraph{Knowledge Editing.}
Knowledge editing \citep{yao2023editing, feng2023trends} is an effective compensatory approach for updating models' knowledge, categorized into four main types: fine-tuning, locate-then-edit, meta-learning, and memory-based methods.
Constrained fine-tuning \citep{meng2022locating}, while straightforward for correcting model behavior, often risks damaging non-edited knowledge.
Locate-then-edit methods typically involve identifying and updating specific parameters, as seen in ROME \citep{meng2022locating}, which uses MLP-based memories for factual edits, and MEMIT \citep{meng2022mass}, which extends this to batch edits. 
Meta-learning approaches, such as MEND \citep{mitchell2021fast}, involve training an external hyper-network to predict updates to the original model. MALMEN \citep{tan2023massive} further addresses the issue of the cancellation effect in MEND by framing parameter shift aggregation as a least-squares problem.
Memory-based methods like SERAC \citep{mitchell2022memory} and GRACE \citep{hartvigsen2024aging} employ working memories to store edits, dynamically selecting parameters based on input similarity. 
Furthermore, recent studies \citep{wang2024wise, wang2024lemoe} have further investigated ways to reduce the adverse effects associated with sequential edits.
While existing approaches predominantly focus on individual models, there is a lack of research addressing the efficient editing of multiple models.

\paragraph{Model Ensemble.}
The model ensemble approach \citep{lu2024merge} integrates the strengths of multiple models to produce refined answers and can be categorized by fusion granularity into output-level and probability-level ensembles \citep{yao2024determine}.
In output-level model ensembles, the outputs from multiple models are combined as candidate sets. 
Methods like PAIRRANKER \citep{jiang2023llm} and routing mechanisms \citep{lu2023routing} select the best candidate based on pairwise comparison or input-specific suitability. 
Other studies \citep{wang2023fusing, jiang2023llm} train fusion modules to integrate outputs effectively.
Probability-level model ensembles, on the other hand, focus on merging the probability distributions of multiple models at each decoding step. 
This process is particularly challenging when dealing with heterogeneous models due to the need for vocabulary alignment. 
To address this issue,  EVA \citep{xu2024bridging} employs overlapping tokens to learn token alignment across different vocabularies, while DEEPEN \citep{huang2024enabling} transforms the representations of each model into a shared space using common vocabulary tokens. 

%% file: section/Conclusion.tex
In this work, we introduce \textsc{OnceEdit} which addresses the challenges of efficiency and stability in multi-model editing scenarios. 
By leveraging a lightweight plug-in model as the editing module and employing improved heterogeneous model ensemble techniques, \textsc{OnceEdit} enables knowledge updates across multiple models with low migration costs.
Extensive evaluations across multiple models and datasets demonstrate that \textsc{OnceEdit} outperforms existing knowledge editing methods in both teacher-forced and validation generation settings. 
Further analysis confirms \textsc{OnceEdit}'s adaptability and stability, underscoring its potential as an effective solution for real-world scenarios.

%% file: section/Limitations.tex
Despite \textsc{OnceEdit}'s high efficiency and adaptability, there are several limitations worth noting.
Firstly, although the plug-in model is relatively small and inference speed can be improved through strategies such as parallel decoding, \textsc{OnceEdit} inevitably incurs additional overhead due to the inclusion of the plug-in model.
Secondly, our experiments focused exclusively on the batch editing setting and did not explore more complex scenarios, such as sequential editing or multi-hop editing tasks.
Thirdly, our study primarily adopted a direct fine-tuning approach for knowledge editing within the plug-in model.
It is important to emphasize that our framework is fundamentally orthogonal to existing knowledge editing methods. 
In future work, we plan to enhance the plug-in model by integrating more advanced editing techniques, enabling us to more effectively address challenges such as sequential editing and generalization.

%% file: section/Acknowledgements.tex
Xiaocheng Feng is the corresponding author of this work. We thank the anonymous reviewers for their insightful comments. This work was supported by the National Natural Science Foundation of China (NSFC) (grant 62276078, U22B2059), the Key R\&D Program of Heilongjiang via grant 2022ZX01A32,  and the Fundamental Research Funds for the Central Universities (Grant No.HIT.OCEF.2023018).

%% file: section/Appendix.tex
\newpage
\section{Implementation Details}
\label{app:main_exp_details}
Our experiment evaluates \textsc{OnceEdit} under batch editing by comparing it with seven knowledge editing methods: FT-L, MEND, ROME, MEMIT, DEFER, GRACE, and WISE. 
All experiments were conducted on NVIDIA A100 80GB GPUs. The hyperparameter settings for these baselines follow previous works \citep{wang2024wise, wang2024lemoe}.
Below, we provide an overview of these methods along with their implementation details:
\begin{itemize}[noitemsep,topsep=0pt,parsep=0pt,leftmargin=*]
\item \textbf{FT-L} is a constrained fine-tuning approach that updates only a single MLP layer while imposing a $L_{\infty}$ norm constraint on weight modifications. 
In our experiments, we select the 21-th layer for GPT-J-6B and the 27-th layer for Llama2-7B and Mistral-7B-v0.1. The fine-tuning learning rate is set to 5e-4.
Notably, to avoid OOM issues, we adopt a batch-based strategy, where we use 5 batches, each updating 200 knowledge entries per round when testing FT-L.
\item \textbf{MEND} is a meta-learning-based approach that trains an external hyper-network to simulate gradients. 
It employs low-rank decomposition with a specialized design to reduce the size of the hyper-network.
For the training phase of MEND, we align the experimental settings entirely with those used in EasyEdit \citep{wang2023easyedit}.
\item \textbf{ROME} locates layers relevant to edits by first disrupting and then restoring activations. 
It subsequently updates the parameters of feedforward networks (FFNs) in a direct manner to modify knowledge.
We select the [3,4,5,6,7,8] layers as the target layer for the GPT-J-6B, and [4,5,6,7,8] for the Llama2-7B and Mistral-7B-v0.1.
\item \textbf{MEMIT} utilizes the same knowledge localization method as ROME but enables simultaneous updates across multiple layers, allowing for the batch integration of thousands of edited knowledge entries.
Consequently, MEMIT and ROME share the same target layer selection.
\item \textbf{DEFER} is a reimplementation of SERAC, utilizes an external memory to store editing instances and trains an additional scope classifier and counterfactual model to appropriately respond to inputs.
we set the learning rate is 7e-5  and select the 21-th layer for GPT-J-6B and the 27-th layer for Llama2-7B and Mistral-7B-v0.1.
\item \textbf{GRACE} leverages a discrete key-value codebook to perform knowledge editing. 
Throughout the editing process, the codebook is dynamically maintained by introducing new keys, expanding existing ones, and splitting them as needed. 
During inference, the method identifies the closest matching key and determines whether to adjust the activation of the hidden layer output.
For the learning rate and $\epsilon_{\text{init}}$, we also align the experimental settings with those used in EasyEdit.
\item \textbf{WISE} stores different edits in separate side memories and routes input queries to the appropriate memory based on activation scores. 
For the hyperparameters setting of WISE, we adhere to the original paper.
\end{itemize}
Apart from the aforementioned methods, there are also two more recent approaches, MEMoe \citep{wang2024memoe} and LEMoE \citep{wang2024lemoe}, whose results are not included due to the unavailability of their source code.

For \textsc{OnceEdit}, we use Tiny-Llama as the plug-in model, setting the learning rate to 1.0e-4 and the multi-task balanced hyperparameter $\lambda=0.8$.
Since both \textsc{OnceEdit} and WISE require the introduction of unrelated knowledge as auxiliary information, we select the instances from the training sets of ZsRE and Counterfact.
Additionally, due to the edited models are the base model, we incorporate prompt-assisted fine-tuning for the ZsRE and Counterfact. The prompts are as follows:
For ZsRE: "Answer this question:\textbackslash n[Question]: \{Input\}\textbackslash n[Answer]:".
For Counterfact: "Complete this half sentence.\textbackslash n [Half Sentence]: \{Input\}\textbackslash n[Answer]:"
To ensure a fair comparison, we evaluate the baselines both with and without prompts, reporting the highest observed value.

\section{Evalution for Knowledge Editing}
\label{appendix:eval_KE}
Following prior research, an effective knowledge editing method should satisfy three essential properties: \textbf{Reliability}, \textbf{Generality}, and \textbf{Locality}. These properties serve as important evaluation metrics for editing methods.

% appendix
\noindent \textbf{Reliability} refers to the model's ability to correctly respond to inputs from the edited set. Specifically, the edited model $\theta^{\prime}$ should consistently produce the correct output for the instances in $S_E$:
\begin{equation}
\mathbb{E}_{(x_e, y_e) \in S_E} \mathbbm{1} \left\{ f_{\theta^{\prime}}\left(x_e \right)=y_e \right\}.
\end{equation}

\noindent \textbf{Generality} refers to the edited model’s capacity to apply the edited knowledge beyond the specific examples in $S_E$. Specifically, the model should be able to correctly respond to the instances in the set $S_R$, where $x_r$ is a rephrased version of an edit $x_e$, and the expected output remains $y_e$:
\begin{equation}
\mathbb{E}_{(x_r, y_e) \in S_R} \mathbbm{1} \left\{ f_{\theta^{\prime}}\left(x_r \right)=y_e \right\}.
\end{equation} 

\noindent \textbf{Locality} emphasizes that the edited model should not alter its behavior on the non-edited knowledge. Specifically, for instances in the dataset $S_L$, which are not affected by the edits, the edited model should produce the same output as before the edit: 
\begin{equation}
\mathbb{E}_{(x_{loc}, y_{loc}) \in S_{L}} \mathbbm {1} \left\{ f_{\theta^{\prime}}\left(x_{loc} \right)= f_{\theta} \left(x_{loc} \right)  \right\}.
\end{equation}

\input{section/tables/generation_table_1000}

\input{section/tables/ablation_lamda_results}

\section{FLOPS of Relative Transfer Matrix}
\label{app:flops}
In this section, we derive the statistical formula for the FLOPS required by \textsc{OnceEdit} to compute the relative transfer matrix during the ensemble stage.
Specifically, given an LLM parametrized by $\theta$, and a tiny model parametrized by $\tilde \theta$, let the vocabulary of LLM be $M$ and that of the small model be $N$.
The dimension of embeddings in LLM is denoted as $d_l$, while that of small model is $d_s$.
Additionally, we define $A$ as the anchor tokens set shared between the two models.

Following DEEPEN \citep{huang2024enabling}, we compute the relative transfer matrices for both the LLM and the tiny model.
Here, we take the LLM matrix $R_l$ as an example, while the derivation for the tiny model follows analogously.
Formally, the relative representation matrix $R_l \in \mathbb{R}^{|M|\times |A|}$ encodes the relative representation of each word $m^{(i)}$ in the LLM's vocabulary. The $i$-th row of $R_l$ is given by:
\begin{equation}
\label{eq:cos}
\begin{aligned}
R_l[i] = (cos(e_{m^{(i)}}, e_{a^{(1)}}),..., cos(e_{m^{(i)}}, e_{a^{(|\mathbb{A}|)}})),
\end{aligned}
\end{equation}
where $e_{m^{(i)}}$ and $e_{a^{(1)}}$ denote the embeddings of word $m^{(i)}$ and $a^{(j)}$, respectively.

To address the representation degeneration of outlier words, DEEPEN applies a softmax normalization to transform the relative representations into a probability distribution:
\begin{equation}
\label{eq:normalized relation representation}
\begin{aligned}
\hat{R_l}[i] = \text{softmax}(R_l[i]).
\end{aligned}
\end{equation}

For Equation \ref{eq:cos}, the cosine similarity between each vocabulary word and the anchor tokens is computed as:
\begin{equation}
\label{eq:R_l}
     R_l[i,j] = \frac{E_l[i] \cdot A_l[j]^T}{\|E_l[i]\| \|A_l[j]\|},
\end{equation}
where $E_l \in \mathbb{E}^{|M|\times |d_l|}$ and $A_l \in \mathbb{E}^{|A|\times |d_l|}$ represents the original word embedding matrix and the anchor words word embedding matrix of the LLM, respectively.

The L2 norm of $E_l$ and $A_l$ is calculated as follows:
\begin{equation}
\label{eq:E_l_L2}
    \|E_l[i]\| = \sqrt{\sum_{j=1}^{d_l} E_[i,j]^2},
\end{equation}
\begin{equation}
\label{eq:A_l_L2}
    \|A_l[i]\| = \sqrt{\sum_{j=1}^{d_l} A_[i,j]^2}.
\end{equation}

Based on Equation \ref{eq:R_l}, \ref{eq:E_l_L2} and \ref{eq:A_l_L2}, the total FLOPS required to calculate $R_l$ consists of three components:the dot product operation, the L2 norm computation, and the final division. These are formulated as follows:
\begin{equation}
\label{eq:dot}
    \text{FLOPS}_{Dot} = |M|*|A|*(d_l+d_l-1),
\end{equation}
\begin{align}
\label{eq:l2}
    \text{FLOPS}_{L2} &= |M|*(d_l+d_l - 1+C_{sqrt}) \notag \\
     &+ |A|*(d_l+d_l - 1+C_{sqrt})+1, 
\end{align}
\begin{equation}
\label{eq:div}
    \text{FLOPS}_{Div} = |M|* |A|,
\end{equation}
where $\text{FLOPS}_{Dot}$, $\text{FLOPS}_{L2}$ and $\text{FLOPS}_{Div}$ correspond to the FLOPS of the dot product, L2 norm, and division operations, respectively. The term $C_{sqrt}$ represents the computational cost of the square root operation, which is a constant.

\input{section/tables/flops_setting}

For the computed $R_l$, a softmax operation is applied for normalization, defined as follows:
\begin{equation}
    \text{softmax}(R_l[i,j]) = \frac{\exp(R_l[i,j])}{\sum_{j=1}^{|A|} \exp(R_l[i,j])}.
\end{equation}

The FLOPS required for the softmax computation on $R_l$ are given by:
\begin{align}
\label{eq:softmax}
    \text{FLOPS}_{softmax} &= |M|*|A|*C_{exp} \notag \\
     &+ |M|*(|A|-1)+|M|*|A|,
\end{align}
where $C_{exp}$ represents the computational cost of the exponential operation, which is a constant.

In summary, the total FLOPS, denoted as $\text{FLOPS}_{all}^l$, required to compute the relative transfer matrix of LLM is the sum of the computational costs from the Equation \ref{eq:R_l}, \ref{eq:E_l_L2} ,\ref{eq:A_l_L2} and \ref{eq:softmax}. For simplification, we set $C_{sqrt} = 2 \,\text{FLOPS}$ and $C_{exp} = 25 \,\text{FLOPS}$. Under this assumption, the $\text{FLOPS}_{all}^l$ is given by:
\begin{align}
\label{eq:all_l}
    \text{FLOPS}_{all}^l &= \text{FLOPS}_{Dot} + \text{FLOPS}_{L2} \notag \\
    & + \text{FLOPS}_{Div} + \text{FLOPS}_{softmax} \notag\\
    & = (2*d_l+27)*|M|*|A| + 1 \notag \\
    & + 2*d_l*|M| + (2*d_l+1)*|A|.
\end{align}
Based on the reasoning process above, We derive the FLOPS equation for the relative transfer matrix of the tiny model:
\begin{align}
\label{eq:all_s}
    \text{FLOPS}_{all}^t &= (2*d_s+27)*|N|*|A| + 1 \notag \\
    & + 2*d_s*|N| + (2*d_s+1)*|A|.
\end{align}

Finally, by integrating Equations \ref{eq:all_l} and \ref{eq:all_s} along with the configurations of each model, as summarized in Table \ref{tab:flops_setting}, we compute the final $\text{FLOPS}_{all}$.

%% file: section/tables/generation_table_1000.tex
\begin{table}[ht]
\centering
\resizebox{\linewidth}{!}{
\begin{tabular}{lcccccccc}
    \toprule
    \multirow{2}{*}{\textbf{Method}}
    & \multicolumn{4}{c}{\textbf{Llama2-7B}} & \multicolumn{4}{c}{\textbf{Mistral-7B-v0.1}} \\
    \cmidrule(lr){2-5} \cmidrule(lr){6-9}
    & Rel.$\uparrow$ & Gen.$\uparrow$ & Loc.$\uparrow$ & Avg.$\uparrow$ 
    & Rel.$\uparrow$ & Gen.$\uparrow$ & Loc.$\uparrow$ & Avg.$\uparrow$ \\
    \midrule
          \multicolumn{9}{c}{\textbf{ZsRE}} \\
    \midrule
    MEMIT   & 0.03& 0.03& 0.00& 0.02& 0.50& \underline{0.48}& 0.04& 0.34 \\
    WISE    & 0.58& \underline{0.46}& \textbf{0.99}& \underline{0.67}& 0.57& 0.46& 0.99& \underline{0.67}  \\
    GRACE   & \textbf{1.00}& 0.00& \textbf{0.99}& 0.66& \textbf{0.98}& 0.03& \textbf{1.00}& \underline{0.67}  \\
    \midrule 
    \rowcolor{blue!15} 
    \textsc{OnceEdit}  & \underline{0.99}& \textbf{0.89}& \textbf{0.99}& \textbf{0.96}& \underline{0.93}& \textbf{0.83}& \underline{0.96}& \textbf{0.91} \\
    \midrule
        \multicolumn{9}{c}{\textbf{Counterfact}} \\
    \midrule
    MEMIT   & 0.03& 0.03& 0.03& 0.03& 0.79& \underline{0.72}& 0.01& 0.50 \\
    WISE    & 0.34& \underline{0.04}& 0.09& 0.16& 0.33& 0.08& \underline{0.65}& 0.35  \\
    GRACE   & \textbf{1.00}& 0.00& \textbf{0.99}& \underline{0.67}& \textbf{0.99}& 0.00& \textbf{0.99}& \underline{0.67}  \\
    \midrule 
    \rowcolor{blue!15} 
    \textsc{OnceEdit}  & \underline{0.99}& \textbf{0.82}& \underline{0.36}& \textbf{0.72}& \underline{0.94}& \textbf{0.76}& 0.37& \textbf{0.69} \\
    \bottomrule 
\end{tabular}
}
\caption{Experimental results on ZsRE and Counterfact under validation generation setting. \textbf{Bold} and \underline{underline} numbers indicate the best and second performance among evaluated methods.}
\label{tab:generation_results}
\end{table}

%% file: section/tables/ablation_lamda_results.tex
\begin{table}[ht]
\centering
\resizebox{\linewidth}{!}{
\begin{tabular}{lcccccccc}
    \toprule
    \multirow{2}{*}{Lambda}
    & \multicolumn{4}{c}{\textbf{Llama2-7B}} & \multicolumn{4}{c}{\textbf{Mistral-7B-v0.1}} \\
    \cmidrule(lr){2-5} \cmidrule(lr){6-9}
    & Rel.$\uparrow$ & Gen.$\uparrow$ & Loc.$\uparrow$ & Avg.$\uparrow$ 
    & Rel.$\uparrow$ & Gen.$\uparrow$ & Loc.$\uparrow$ & Avg.$\uparrow$ \\
    \midrule
          \multicolumn{9}{c}{\textbf{ZsRE}} \\
    \midrule
    0.00& 0.05& 0.04& 0.99& 0.36& 0.05& 0.05& 0.98& 0.36 \\
    0.20& 0.99& 0.95& 0.99& 0.98& 0.94& 0.90& 0.98& 0.94 \\
    0.40& 0.99& 0.97& 1.00& \textbf{0.99}& 0.94& 0.92& 0.99& \textbf{0.95} \\
    0.60& 0.99& 0.95& 0.99& 0.98& 0.94& 0.91& 0.98& 0.94 \\
    0.80& 0.99& 0.98& 1.00& \textbf{0.99}& 0.93& 0.93& 0.98& \textbf{0.95} \\
    1.00& 0.99& 0.97& 1.00& \textbf{0.99}& 0.94& 0.91& 0.99& \textbf{0.95}\\
    \midrule
        \multicolumn{9}{c}{\textbf{Counterfact}} \\
    \midrule
    0.00& 0.00& 0.00& 0.99& 0.33& 0.01& 0.00& 0.96& 0.32 \\
    0.20& 0.99& 0.84& 0.68& \textbf{0.83}& 0.94& 0.80& 0.67& \textbf{0.80} \\
    0.40& 0.99& 0.70& 0.76& 0.82& 0.95& 0.72& 0.66& 0.78 \\
    0.60& 0.99& 0.80& 0.71& \textbf{0.83}& 0.95& 0.77& 0.70& \textbf{0.80} \\
    0.80& 0.99& 0.89& 0.53& 0.80& 0.95& 0.85& 0.49& 0.76 \\
    1.00& 0.95& 0.84& 0.67& \textbf{0.83}& 0.95& 0.80& 0.66& \textbf{0.80} \\
    \bottomrule 
\end{tabular}
}
\caption{Ablation study on the impact of lambda ($\lambda$) during the editing stage, evaluated on 200 edited instances under validation generation.
\textbf{Bold} numbers indicate the best performance among diverse settings.}
\label{tab:lamda_ablation}
\end{table}

%% file: section/tables/flops_setting.tex
\begin{table}[t!]
\centering
\resizebox{\columnwidth}{!}{
\begin{tabular}{lccc}
\toprule
     Models &  Vocab. Size& AW Num& Embedding Dim.\\
     \midrule
     Llama2-7B &  32000& 31999&  4096 \\
     Mistral-7B-v0.1 &  32000& 24184&  4096\\
     GPT-J-6B &  50400& 17830& 4096\\
     Tiny-Llama & 32001& - &2048 \\
     \bottomrule
\end{tabular}
}
\caption{configurations for calculating relative transfer matrix FLOPS.
Vocab. Size denotes the vocabulary size, AW Num represents the number of anchor words shared between the models and Tiny-Llama, and Embedding Dim. refers to the dimension of the embedding layer.}
\label{tab:flops_setting}
\end{table}